\documentclass{article}
\usepackage{arxiv}
\pdfoutput=1
\usepackage{times}
\usepackage{soul}
\usepackage{url}
\usepackage[utf8]{inputenc}
\usepackage[small]{caption}   %?????????????????????????????
\usepackage{graphicx}
\usepackage{amsmath}
\usepackage{amsthm}
\usepackage{booktabs}
\usepackage{algorithm}
\usepackage{algorithmic}

\usepackage{threeparttable}
\usepackage{tabularx}
\usepackage{enumerate}
\usepackage{bbm}
\usepackage{amsfonts}
\usepackage{epsfig, epstopdf}
\usepackage{subfigure}
\usepackage{multirow}
\usepackage{xcolor}

\usepackage{amssymb}
\usepackage{flushend}

\usepackage{graphicx}
\usepackage{subfigure}
\usepackage{amsmath,amsfonts,amssymb}
\usepackage{lineno}
\usepackage{siunitx}

\usepackage{indentfirst}
\usepackage[T1]{fontenc}
\usepackage[utf8]{inputenc}
\usepackage{authblk}

\usepackage[utf8]{inputenc}
\usepackage{graphicx}
\providecommand{\keywords}[1]{\textbf{\textit{Index terms---}} #1}

\title{Deep Embedded Multi-view Clustering with Collaborative Training}

\author[a]{Jie Xu}
\author[{a,$*$}]{Yazhou Ren}
\author[a]{Guofeng Li}
\author[a]{Lili Pan}
\author[a]{Ce Zhu}
\author[b,c]{Zenglin Xu}
\affil[a]{University of Electronic Science and Technology of China, Chengdu, China}
\affil[b]{Harbin Institute of Technology, Shenzhen, China}
\affil[c]{Peng Cheng Lab, Shenzhen, China}

\date{}

\begin{document}
%-------------------------------------------------------------------------
\maketitle

% \footnote{Corresponding author: yazhou.ren@uestc.edu.cn}
\let\thefootnote\relax\footnotetext{$^{*}$Corresponding author: yazhou.ren@uestc.edu.cn} 

\begin{abstract}
Multi-view clustering has attracted increasing attentions recently by utilizing information from multiple views. However, existing multi-view clustering methods are either with high computation and space complexities, or lack of representation capability. To address these issues, we propose \underline{d}eep \underline{e}mbedded \underline{m}ulti-\underline{v}iew \underline{c}lustering with collaborative training (DEMVC) in this paper. Firstly, the embedded representations of multiple views are learned individually by deep autoencoders. Then, both consensus and complementary of multiple views are taken into account and a novel collaborative training scheme is proposed. Concretely, the feature representations and cluster assignments of all views are learned collaboratively. A new consistency strategy for cluster centers initialization is further developed to improve the multi-view clustering performance with collaborative training. Experimental results on several popular multi-view datasets show that DEMVC achieves significant improvements over state-of-the-art methods. The code and datasets are available at \url{https://github.com/JieXuUESTC/DEMVC}.
%\url{https://t.cn/A6vkn5Rk}.
\end{abstract}

% \begin{graphicalabstract}
% \includegraphics[height=3.5in, width=6.22in]{pic/1.pdf}
% \end{graphicalabstract}

% \begin{highlights}
% \item A novel deep embedded multi-view clustering method is proposed, which can well utilize the common and complementary information of multiple views by training multiple deep neural networks collaboratively.
% \item A shared scheme of the auxiliary distribution and a new consistency strategy of cluster centers initialization are developed to improve the performance of multi-view clustering.
% \item The proposed model has good representation capability. In addition, it can be solved efficiently and applied to large-scale datasets. Experiments on several popular datasets demonstrate that DEMVC achieves state-of-the-art performance.
% \end{highlights}

% \begin{keywords}
% Deep embedded clustering \sep Multi-view clustering \sep Unsupervised learning \sep Collaborative training
% \end{keywords}

\section{Introduction}
Cluster analysis is a fundamental unsupervised learning task in machine learning, which categorizes data samples without labels based on their association with each other~\cite{macqueen1967some,ng2002spectral,zhong2020nonnegative,abdolali2020neither}. Recently, clustering methods based on deep neural networks (DNN) have achieved impressive clustering performance \cite{xie2016unsupervised,ghasedi2017deep,yang2017towards,chang2017deep,ren2018deep,mrabah2019adversarial}. However, these methods typically solve a single-view clustering problem.

%\cite{zhong2020nonnegative} proposed a subspace clustering with an adaptive affinity matrix learning method. Besides, based on an augmented Lagrangian multiplier, an efficient iterative algorithm was proposed to optimize this subspace clustering method;

In many real-world clustering tasks, a data example often has different observable views. For instance, a webpage can be described by both page content and linkage, an object can be captured with different poses, a handwritten digit can be written by different persons.
To significantly make use of multiple views’ complementary information to enhance the clustering performance, multi-view clustering (MVC) has been proposed. The consensus and complementary principles are two basic concepts in multi-view clustering \cite{xu2013survey}. 
On the one hand, since multiple views are exactly multiple maps of the same object, the consensus principle seeks to make multiple predictions of the same object consistent among multiple views. On the other hand, due to the diversity of different views, the complementary principle aims at comprehensively utilizing the complementary information of all views to make better predictions.

%The biggest difference between multi-view clustering and single-view clustering is that the objects to be processed by multi-view clustering have multiple operable subsamples, which are different observations of the same object. Therefore, t
The key of multi-view clustering is to effectively mine the information contained in multiple views to achieve better clustering performance. Multi-view information includes common information and complementary information among multiple views. Common information refers to the similar information contained in multiple views. For example, both of two pictures about cats have contour and facial features. The common information of multiple views is helpful to improve the understanding of the commonness of the research objects. Complementary information means that multiple views have specific information about the same object. One view often contains incomplete information, while complementary information can complement each other. For example, one view shows the side of a cat and the other shows the front of the cat, these two views allow for a more complete depiction of the cat. In order to fully depict the cat, all these scattered complementary information in different views is useful. 
%But for multi-view clustering, specific information does not necessarily represent complementary information and may be redundant information that has interference to clustering. The purpose of multi-view learning is to mine both common and complementary information while eliminate the influence of interference information. 
In our study, extracting common information and complementary information are corresponding to the consensus and complementary principles, respectively. 
%In fact, single-view clustering can be considered as a special case of multi-view clustering.

In general, the multi-view clustering approaches can be divided into four categories as below: 
(1) canonical correlation analysis based MVC, e.g., \cite{andrew2013deep,wang2015deep}, associates two related views to explore information that is conducive to clustering;
(2) subspace clustering based MVC, e.g., \cite{cao2015diversity,abavisani2018deep}, explores a shared representation of multiple views to obtain a similarity metric matrix for spectral clustering; 
(3) matrix factorization based MVC, e.g., \cite{zhao2017multi,zhang2018binary}, decomposes each view into a low-rank matrix with specific constraints and then applies a specific clustering algorithm;
(4) graph based MVC, e.g., \cite{nie2017self,peng2019comic}, uses multiple views' information to construct graphs for clustering.

Although existing MVC methods have been successfully applied in various fields, they still have two main disadvantages. (1) Traditional shallow MVC approaches have limited representation capability and are not applicable for many applications, e.g., image clustering. (2) Existing methods usually solve spectral clustering or matrix factorization problems, leading to high computation and space complexities. Thus, these methods can not handle large-scale data clustering tasks. 

To address the above mentioned issues, we propose a novel MVC model in this work, namely deep embedded multi-view clustering with collaborative training (DEMVC).
%We were inspired by the movement of climbing stairs: lifting the body step by step alternately on both legs, with the higher leg pulling on the lower leg. Similarly, 
Considering a simple clustering of two views, we set one view as the referred view and use its objective to guide the training of itself and the other view. We expect that the view with better clustering performance will serve as a guide to the other view, so as to better mine the common information and complementary information in both views that is beneficial to clustering and enhance the clustering performance. This idea can be extended to clustering of multiple views.

To this end, DEMVC firstly performs feature learning individually for each view by employing deep autoencoders --- with good representation capability --- to obtain the embedded feature representations. Secondly, DEMVC applies $k$-means on one view (which is named the referred view) to obtain the initial cluster centers in the embedded space and then computes the auxiliary target distribution of this view. 
This auxiliary distribution of the referred view is used to refine the deep autoencoders and clustering soft assignments for all views. Each view will become the referred view in sequence to ensure that the multi-view clustering takes full advantage of all views. Performing such collaborative training, all views share the same auxiliary target distribution in every round, and each view can learn from its own view and the other views. Finally, the clustering assignments of all views are summarized to generate the final clustering result. It is verified that the consensus and complementary principles of MVC are guaranteed by the proposed framework.
%These information will be used to refine the deep autoencoders and cluster assignments of all the views. Each view will become such referred view in turn. In this way, both the consensus and complementary principles of MVC are guaranteed because the same auxiliary distribution are shared across all views when refining, and each view can learn from its own data and from the information of other views simultaneously. 

%So in this paper, based on the ability of the autoencoder to reconstruct samples and the ability of deep clustering to characterize mass of data, we proposed a novel multiple networks collaborative training method to implement multi-view clustering. Under this framework, we explore the role of collaborative training and propose a new strategy to initialize the cluster centers, which further improve its clustering performance. The proposed method in this paper take turns using multi-view information for training, which makes multi-view predictions for samples tend to be similar. So it follows both consensus and complementary principles. 

In summary, the contributions of this paper are three-folds:
\begin{itemize}
\item We propose a novel deep embedded multi-view clustering method, which can well utilize the common and complementary information of multiple views by training multiple deep neural networks collaboratively.
\item A shared scheme of the auxiliary distribution and a new consistency strategy of cluster centers initialization are developed to improve the performance of MVC.
\item The proposed model has good representation capability. In addition, it can be solved efficiently and applied to large-scale datasets. Experiments on several popular datasets demonstrate that DEMVC achieves state-of-the-art performance. 
\end{itemize}

\section{Related Work}
\label{sec:related}

\textbf{Deep clustering}. In recent years, a number of clustering methods based on deep neural networks have been proposed. 
In deep embedded clustering (DEC) \cite{xie2016unsupervised}, the cluster assignment and the deep autoencoders are jointly learned.
To avoid distortion of the embedded space, \cite{guo2017improved} proposed an improved version of DEC (IDEC).
\cite{guo2017deep} proposed a deep convolutional embedded clustering algorithm to improve the performance of DEC on image data. Specifically, it uses a convolutional autoencoder to learn the embedded feature space in an end-to-end manner.
\cite{ghasedi2017deep} further improved the performance of DEC by stacking multinomial logistic regression function on top of a multi-layer convolutional autoencoder.
\cite{ji2017deep} introduced a deep neural network with a novel self-expressive layer to improve the traditional subspace clustering.
\cite{zhang2018scalable} introduced a method that uses deep neural networks to update multiple subspaces and simultaneously combine the reconstruction error to obtain the embedded space, which realizes end-to-end subspace clustering.
\cite{ren2019semi} applied DEC to semi-supervised clustering.
\cite{zhang2019neural} proposed a cooperative subspace clustering method, which can find clusters of data points from a joint low-dimensional subspace. 
%Specifically, the supervised collaboration scheme is implemented through two affinity matrices, one from the concept of self-expression in the subspace clustering and the other from the classifier.

In this paper, we follow the idea of deep embedded clustering and use the same deep autoencoder as \cite{guo2017improved,guo2018deep,ren2018deep}, to learn the representations and clustering assignments of samples in low-dimensional embedded space for each view. In addition, we propose a multi-view collaborative training strategy to apply deep embedded clustering to multi-view learning.

%Generative Adversarial Network(GAN) is also used in clustering problems, such as %Info-GAN (-----------) and ClusterGAN (---------). They generate and cluster %images and enforce the discriminative feature in latent space.

\textbf{Multi-view clustering}. 
Canonical correlation analysis (CCA) \cite{anderson1958introduction} is used to find linear projections of two maximally correlated random vectors.
\cite{andrew2013deep} explored linearly correlated representation by learning nonlinear transformations of two views with deep canonical correlation analysis (DCCA). 
\cite{wang2015deep} proposed deep canonically correlated autoencoders (DCCAE), which is an improved version of DCCA. 
Subspace multi-view clustering approaches assume that multiple views of data come from the same latent space.
\cite{cao2015diversity} proposed a diversity-induced multi-view clustering method by extending the traditional subspace clustering. 
\cite{zhang2017latent} proposed a multi-view clustering method by learning the potential subspace representations of samples.
%\cite{ji2017deep} introduced a deep neural network with a novel self-expressive layer to improve the traditional subspace clustering.
\cite{abavisani2018deep} applied deep learning to multi-modal subspace clustering.
\cite{brbic2018multi} presented a multi-view sparse subspace clustering method.
Combining with convolutional autoencoders and CCA-based self-expressive module, \cite{tang2018deep} introduced a deep multi-view sparse subspace clustering.
\cite{zhang2018binary} proposed to encode collaboratively descriptors of multi-view images into a binary code space. %, which can easily scale to large data.
\cite{lin2018jointly} used different kinds of autoencoders to learn multiple deep embedded features and clustering assignments with multi-view fusion mechanism. 
\cite{peng2019comic} proposed a multi-view clustering method to firstly learn a connection graph for each view and then minimize the discrepancy of pairwise connection graphs.
\cite{nie2017self} explored Laplacian rank constrained graph and proposed a self weighted multi-view clustering method.
\cite{huang2020auto} proposed a novel multi-view co-clustering method, which learns optimal weight for bipartite graphs automatically.
\cite{yu2020active} proposed a three-way multi-view data clustering (uncertain, belong-to and not belong-to) via low-rank matrices.
\cite{xu2015multi,ren2019self} applied self-paced learning in multi-view clustering to address the non-convexity issue.

Based on the consensus principle and complementary principle of multi-view clustering, this paper proposes a novel deep embedded multi-view clustering method. Unlike the above mentioned MVC methods, our approach achieves the consistency of multi-view prediction by a novel collaborative training strategy which shares an auxiliary distribution alternately. Experiments show that this method can mine both the common information and complementary information in multi-view data, and can improve the multi-view clustering performance significantly.

\section{Proposed Method}\label{sec:proposed method}
This section presents our deep embedded multi-view clustering  with  collaborative  training  (DEMVC) in detail.
\subsection{Multi-view Collaborative Training}
\label{sec:Multi-network}

Consider clustering the dataset $\{\pmb{x}^v_i\in \mathbb{R}^D\}_{i=1}^N$ into $K$ clusters, where $N$ is the number of samples and $D$ is the dimensionality.
Let $V$ be the number of views and represent that there are $V$ subsamples for each object to be clustered. For the $v$-th view ($v=1,2,\ldots,V$), let $f_{\Theta}^v$ and $g_{\Omega}^v$ be the encoder and decoder, respectively.  $\Theta$ and $\Omega$ are the corresponding learnable parameters.
Based on the nonlinear mapping ability of neural networks, $f_{\Theta}^v$ and $g_{\Omega}^v$ realize:

\begin{equation}\label{eq:encoder}
    \pmb{z}_i^v=f^v_{\Theta}{(\pmb{x}_i^v)}
\end{equation}
and
\begin{equation}\label{eq:decoder}
    \hat{\pmb{x}}_i^v=g^v_{\Omega}{(\pmb{z}_i^v)}=g^v_{\Omega}{(f^v_{\Theta}{(\pmb{x}_i^v)})},
\end{equation}

\noindent where $\pmb{z}_i^v \in \mathbb{R}^d$ is the embedded point, which is encoded by $f_{\Theta}^v$, of $\pmb{x}_i^v$ in the low $d$-dimensional feature space. %, which dimensionality is $d$. 
$g_{\Omega}^v$ decodes $\pmb{z}_i^v$ and reconstructs the sample as $\hat{\pmb{x}}_i^v \in \mathbb{R}^D$. For multi-view clustering, we define the loss function as:

\begin{equation}\label{eq:Loss}
    L=\sum_{v=1}^V L_r^v+\gamma\sum_{v=1}^V L_c^v,
\end{equation}

\noindent where $L_r^v$ and $L_c^v$ are the reconstruction loss and clustering loss of the $v$-th view, respectively. $\gamma$ is a trade-off coefficient. %the coefficient that balances the two types of losses. 
In fact, Eq.~(\ref{eq:Loss}) can be considered as a multi-view generalized version of IDEC \cite{guo2017improved}. Following IDEC, the trade-off parameter $\gamma$ is set to 0.1. 
%Following the single-view deep embedded clustering~\cite{guo2017improved}, the trade-off parameter $\gamma$ between $L_r^v$ and $L_c^v$ is set to 0.1. 
Combined with Eq.~(\ref{eq:decoder}), the reconstruction loss is defined as:

\begin{equation}\label{eq:Lr}
    L_r^v=\frac{1}{N}\sum_{i=1}^N
    \left\|
    {\pmb{x}_i^v-g^v_{\Omega}{(f^v_{\Theta}{(\pmb{x}_i^v)})}}
    \right\|
    _2^2.
\end{equation}

Define $\pmb{\mu}_j^v \in \mathbb{R}^d$ as the $j$-th cluster center of the $v$-th view. The Kullback-Leibler (KL) divergence is used to define the clustering loss:

\begin{equation}\label{eq:Lc}
    L_c^v=KL(P^v||Q^v)=\sum_{i=1}^N\sum_{j=1}^K p_{ij}^v\log{\frac{p_{ij}^v}{q_{ij}^v}},
\end{equation}

\noindent where $q_{ij}^v\in \mathbb{R}^K$ is the similarity between the embedded point $\pmb{z}_i^v$ and cluster center $\pmb{\mu}_j^v$ and is calculated by Student's $t$-distribution~\cite{maaten2008visualizing} as: 

\begin{equation}\label{eq:q}   
    q_{ij}^v=\frac{(1+\lVert \pmb{z}_i^v-\pmb{\mu}_j^v\rVert^2)^{-1}}
    {\sum_{j}(1+\lVert \pmb{z}_i^v-\pmb{\mu}_{j}^v\rVert^2)^{-1}}. 
\end{equation}

\noindent In a clustering task, $q_{ij}^v$ is treated as the soft label that represents the probability of assigning the $i$-th sample of the $v$-th view to the $j$-th category. With the operation of square and normalization of clustering soft label $q_{ij}$,
%By raising $q_{ij}$ to the second power, 
DEC~\cite{xie2016unsupervised} built the auxiliary target $p_{ij}\in \mathbb{R}^K$ to implement deep single-view clustering. 
%By minimizing the clustering loss, the distribution of clustering soft labels $Q$ will approach to the auxiliary target distribution $P$, and a clearer clustering structure can be obtained. So
Similarly, $p_{ij}^v$ in our deep multi-view clustering is calculated by:

\begin{equation}\label{eq:p}        
    p_{ij}^v=\frac{{(q_{ij}^v)^2}/{\sum_i q_{ij}^v}}
    {\sum_{j}({(q_{i{j}}^v)^2}/{\sum_i q_{i{j}}^v})}.
\end{equation}

\begin{figure}
\centering
\subfigure[Before optimization]{\includegraphics[height=1.5in, width=1.5in]{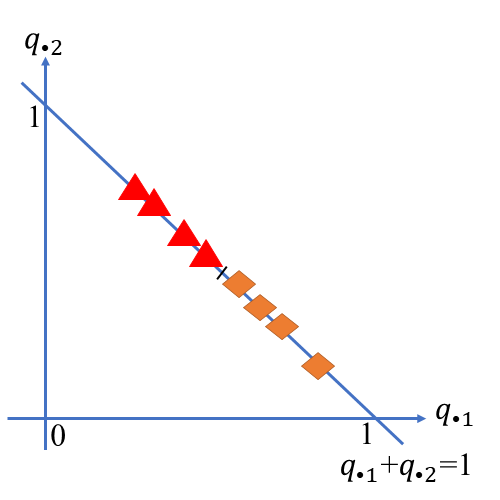}}
\subfigure[After optimization]{\includegraphics[height=1.5in, width=1.5in]{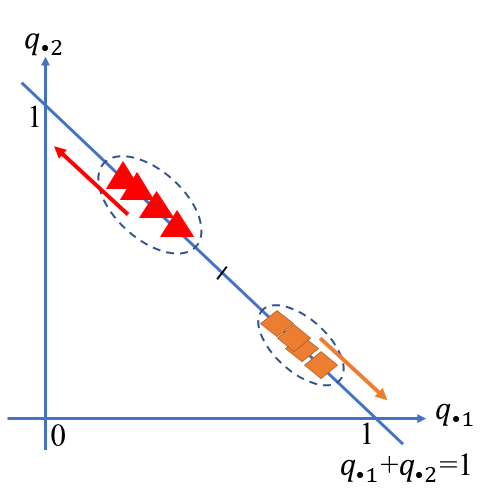}}
\subfigure[Before optimization]{\includegraphics[height=1.5in, width=1.5in]{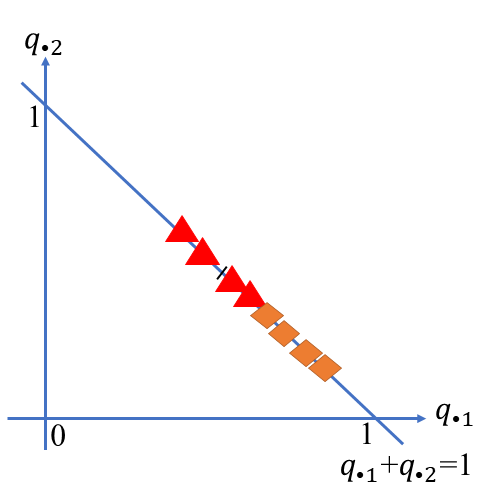}}
\subfigure[After optimization]{\includegraphics[height=1.5in, width=1.5in]{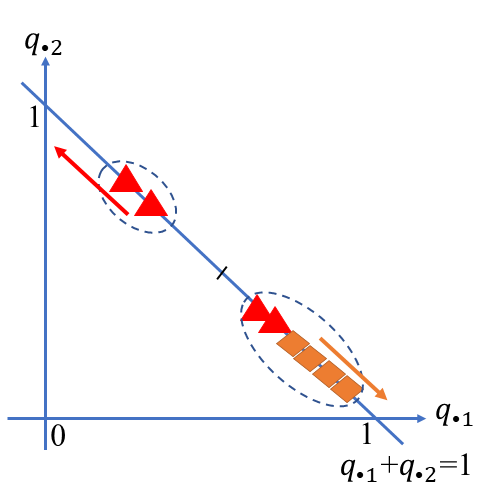}}
\caption{The mechanism of deep embedded clustering to minimize KL divergence. Triangles and rhomboids represent different classes.The two coordinates ($q_{\bullet 1}$ and $q_{\bullet 2}$) of a point represent the probabilities that the corresponding sample belongs to two classes, respectively. Obviously, $q_{\bullet 1}+q_{\bullet 2}=1$ holds.}
\label{fig:IDEC}
\end{figure}

%Through the auxiliary target $p$, the optimization target of the $K$-class sample is to get the point coordinate (..., 0, ..., 1, ..., 0, ...). 
%The autoencoders have representation capability, but they don't know which characteristic are useful information for clustering. 
The autoencoders have good representation capability, but the learned representations may not suitable for clustering.
We regard the soft label $q$ as a point in the $K$-dimensional space. 
By minimizing the KL divergence of $Q$ and $P$, DEC and IDEC refine autoencoders such that the soft label $q$ is more distinguishing. We take 2-class clustering for example as shown in Figure \ref{fig:IDEC}(a) and (b). After minimizing the KL divergence, the two clusters become better separated. 
%When DEC and IDEC minimize the KL divergence of $Q$ and $P$, they encourage autoencoders to learn network weights that make $q$ more precise, we take 2-class clustering for example as shown in Figure \ref{fig:IDEC}(a) and (b). 
Therefore, the refined encoders can acquire more discriminative clustering capabilities, this is why DEC and IDEC can achieve impressive clustering performance. However, they optimize the KL divergence according to a single view. In this way, those hard samples (the samples that are close to other classes, that are fuzzy and hard to distinguish) are prone to misclassification. 
%Which are in the middle of the coordinate space 
As shown in Figure \ref{fig:IDEC}(c) and (d), some samples which near the classification boundary might be misclassified.

If Eq.~(\ref{eq:Loss}) is optimized directly, the embedded features and clustering assignments of each view will be learned independently. Thus, the complementary information of multiple views is ignored. 
But, as we mentioned previously, more complementary information should be used and the clustering prediction of multiple views should be consistent as far as possible. Hence, in order to use the common information and complementary information of multiple views, we let each view become the referred view in turn to guide the whole networks to learn the features, which are conducive to clustering. This training idea is called multi-view collaborative training. Specifically, we define $P^r$ as the auxiliary target distribution of the referred view. In Eq.~(\ref{eq:Lc}), let $P^r$ be the shared auxiliary target distribution for all views. Then, the clustering loss of the $v$-th view is:

\begin{equation}\label{eq:Lc_iter}
    L_c^v=KL(P^r||Q^v).
\end{equation}

\begin{figure}
\centering
\subfigure[Before coordinate alignment]{\includegraphics[height=1.5in, width=1.5in]{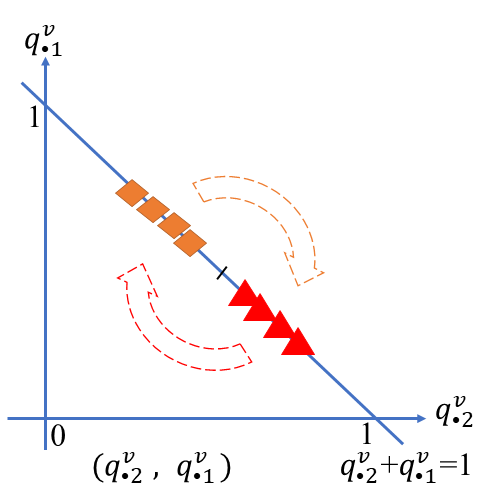}}
\subfigure[After coordinate alignment]{\includegraphics[height=1.5in, width=1.5in]{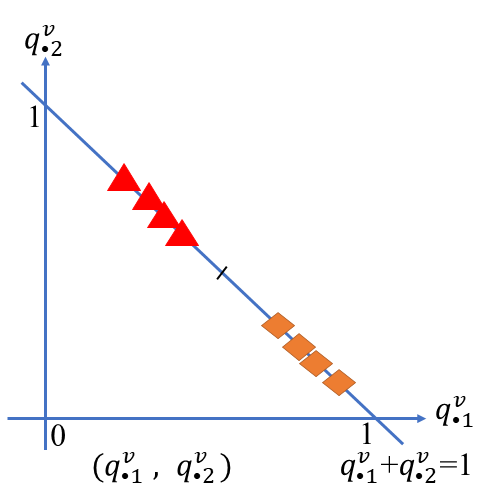}}
\subfigure[The referred view with correct samples]{\includegraphics[height=1.5in, width=1.5in]{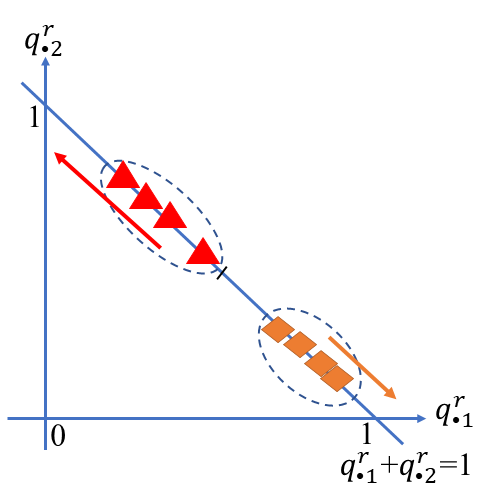}}
\subfigure[Another view with mispredicted samples]{\includegraphics[height=1.5in, width=1.5in]{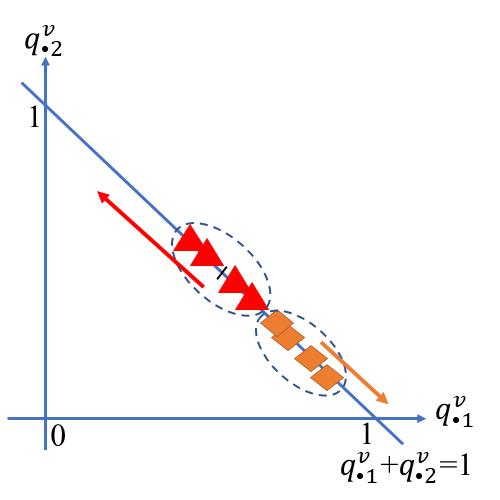}}
\caption{The mechanism by which DEMVC works.}
\label{fig:CT}
\end{figure}

When a view becomes the referred view, all views are collaboratively trained for a certain amount of iterations. Let ($q_{\bullet 1}^r$, $q_{\bullet 2}^r$) represent the referred view's coordinates, as shown in Figure \ref{fig:CT}(a) and (b). Since the proposed collaborative training shares the same auxiliary target distribution $P^r$ across all views, it can align the coordinates of the other views according to the referred view's coordinate system. When setting the referred view in multiple views, the accurately predicted samples of the referred view can correct the mispredicted samples in other views, as shown in Figure \ref{fig:CT}(c) and (d).
%based on Mini-Batch Gradient Descent optimization algorithm
However, the referred view may also have some mispredicted samples which may mislead other views. 
%in the middle of the coordinate space, so if we update all the views by one fixed referred view, it's the same as IDEC
Therefore, it is necessary to change the referred view alternately during collaborative training, so that multiple views can supervise each other to obtain more accurate $Q^v$. In this way, every view can become the referred view in turn to mine the useful complementary information of multiple views to obtain better clustering performance. This corresponds to the complementary principle of multi-view clustering.
%Better samples with more accurate $q$ play a bigger role in multi-view mutual supervision.

%\textcolor{blue}{Multi-view samples contain more useful information than single-view samples. The useful information reflect the essence of samples and can guide to correct clustering. In extreme cases, there are only one view is correct among multiple views about a sample. The information contained in the correct view is critical for clustering. Through multi-view collaborative training, every view can become the referred view in turn to mine the useful complementary information of multiple views, which makes the sample clustering more fault tolerant. This is the embodiment of multi-view clustering complementary principle, so the performance of DEMVC is much higher than IDEC.}

After incorporating Eq.~(\ref{eq:Lc_iter}), the new clustering loss function of collaborative training, Eq.~(\ref{eq:Loss}) becomes:

\begin{equation}\label{eq:LOSS}
\begin{aligned}
    L=\sum_{v=1}^V\frac{1}{N} \sum_{i=1}^N\left\|{\pmb{x}_i^v-g^v_{\Omega}{(f^v_{\Theta}{(\pmb{x}_i^v)})}}\right\|_2^2 + \gamma\sum_{v=1}^V KL(P^r||Q^v).  
\end{aligned}
\end{equation}

%the auxiliary target $p_{ij}^v$ is the operation of square and normalization of clustering soft label $q_{ij}^v$, under the guidance of clustering soft label distribution $Q$, 
%\textcolor{blue}{Since minimizing the KL divergence of $Q^v$ and $P^v$ will make the probability of $q_{ij}^v$ representing clustering to a certain class increase. In Eq.~(\ref{eq:q}), based on Student's $t$-distribution, if the clustering center $\pmb{\mu}_j^v$ is initialized by a simple clustering algorithm -- such as $k$-means clustering~\cite{macqueen1967some}, then optimization of clustering loss can enable the autoencoders to learn a clearer and better clustering structure compared with the initialization time. Driven by clustering optimization, we take clustering loss as the second term in Eq.~(\ref{eq:LOSS}).}

%For clustering purpose, the optimization of clustering loss, as the second term of Eq.~(\ref{eq:LOSS}), is going to enable the networks to learn a clearer and better clustering structure when the clustering centers is initialized by a simple clustering algorithm -- such as $k$-means clustering~\cite{macqueen1967some}.
%which is guided by the results of $k$-means clustering in the initialization stage. 

For the collaborative training --- where multiple views in turn become the referred view, it is necessary to optimize the reconstruction loss, as the first term of Eq.~(\ref{eq:LOSS}), to maintain the representation capability of the autoencoders for all views. Otherwise, because of the differences between views, the referred view tend to destroy the capability of the other autoencoders to extract the information from their views. In this way, the autoencoders cannot accurately extract the common information and complementary information of multiple views, resulting in poor clustering performance. Therefore, both clustering loss and reconstruction loss are retained in DEMVC.

\begin{figure}
\centering
\subfigure{\includegraphics[height=3.5in, width=6.22in]{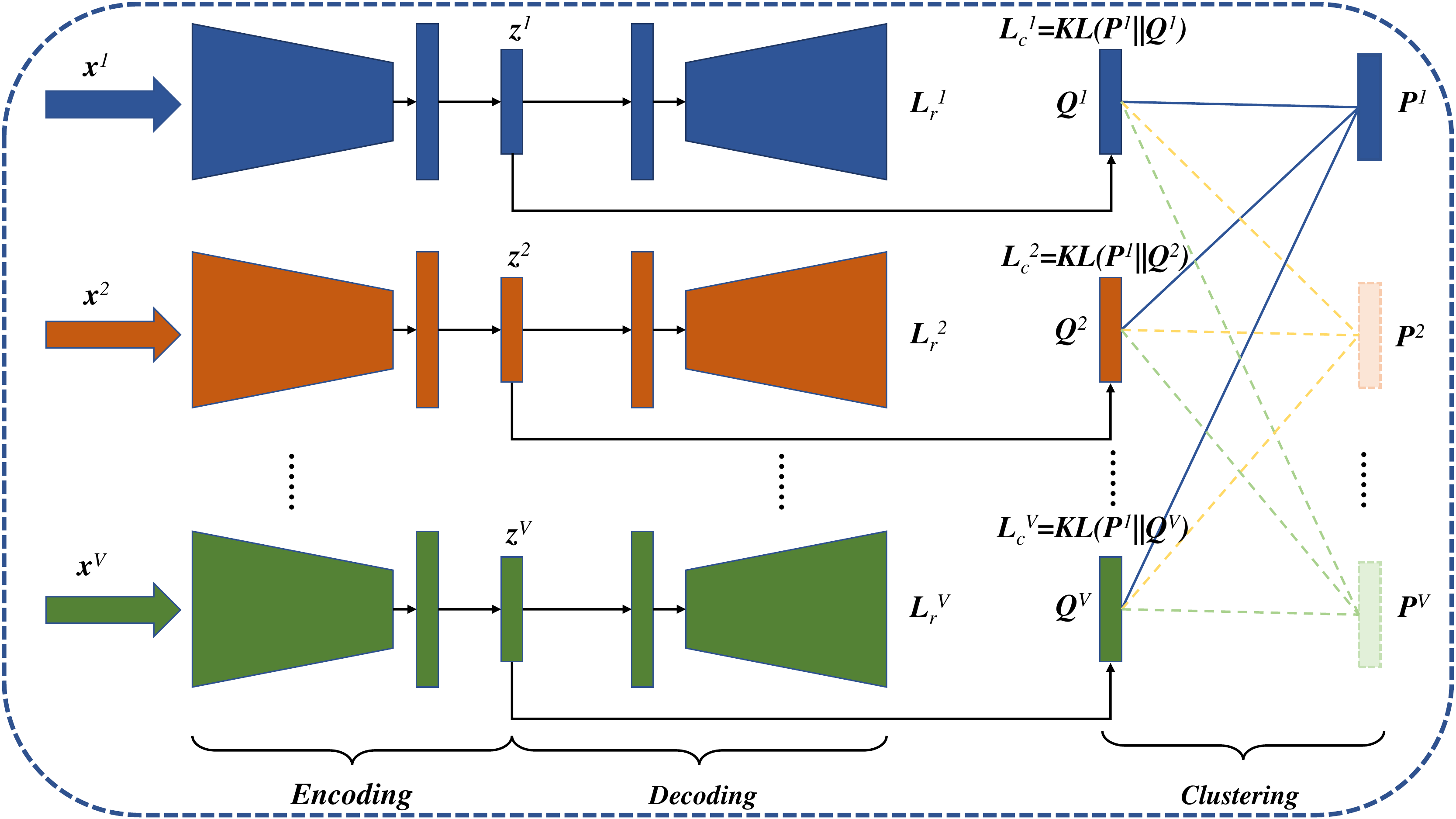}}
\caption{The framework of DEMVC. It consists of $V$ encoders, $V$ decoders, and $V$ clustering layers. The diagram shows that the first view is becoming the referred view as shown with solid lines. The dashed lines represent the corresponding $P^i$s do not participate in calculation in this round. The referred view will change sequentially.}
\label{fig:framework}
\end{figure}

%Before the initialization phase, the encoders do not have the ability to extract the features of the input data. 
The autoencoders, in the beginning, with random network parameters do not have representation capability of the input data. In order to avoid that the referred view guides blindly the training of all views, we first pre-train deep autoencoders of all views via minimizing Eq.~(\ref{eq:Lr}). Then, we collaboratively fine-tune all autoencoders, cluster assignments and cluster centers of all views by minimizing Eq.~(\ref{eq:LOSS}). Please refer to Section \ref{sec:Details} for implementation details. The training framework of our model is shown in Figure \ref{fig:framework}.

When the fine-tuning phase is finished, based on the soft label $q_{ij}^v$, the clustering prediction of $i$-th sample of $v$-th view is computed by:
\begin{equation}\label{eq:sv}
    s_i^v=\arg\mathop{\max_j}{(q_{ij}^v)}.
\end{equation}

%\textcolor{blue}{Most clustering objects have consistent multi-view clustering soft label after collaborative training, that is, the $q_{ij}^v$ of multiple views are aligned, which is the embodiment of multi-view clustering consensus principle.}
In fact, with the influence of shared auxiliary target distribution $P^r$, the final status of collaborative training is that the distribution of clustering soft labels $Q^v$ of multiple views are similar to each other. For most samples, the predictive soft labels $q_{ij}^v$ of their multiple views are consistent and aligned, corresponding to the consensus principle of multi-view clustering. The final prediction is obtained by averaging the multi-view clustering soft labels as:

\begin{equation}\label{eq:s}
    s_i=\arg\mathop{\max_j}{(\frac{1}{V}\sum_{v=1}^V q_{ij}^v)}.
\end{equation}
% the upper bound of the accuracy of multiple views prediction $\pmb{s}^v$ is the lower bound of the accuracy of the final predictive $\pmb{s}$.

\noindent According to Eq. (\ref{eq:s}), for the samples with consistent prediction of multiple views, the final prediction is the same. For those samples whose predictions are inconsistent, the final prediction have the chance to correct the misaligned prediction by integrating the soft labels of multiple views.

\subsection{Consistency strategy for cluster centers initialization}
\label{sec:Cluster center}

According to Eq.~(\ref{eq:q}), $q_{ij}^v$ is calculated by $\pmb{z}_{i}^v$ and $\pmb{\mu}_j^v$. Before the fine-tuning phase, consider setting the cluster centers $\pmb{\mu}_j^v$ of mutiple views to be the same to better follow the consensus principle. In this way, multiple views are not limited to their own cluster centers and are easier to accept the guidance of the referred view. We use $k$-means~\cite{macqueen1967some} to initialize the clustering centers in the first referred view (denoted by $V_s$). The corresponding loss function is:

\begin{equation}\label{eq:kmeans}
    L_{k-means}^{V_s}=\sum_{i=1}^N\sum_{j=1}^K \left\| {\pmb{z}_i^{V_s}-\pmb{c}_j^{V_s}} \right\| ^2,
\end{equation}

\noindent where $\pmb{c}_j^{V_s} \in \mathbb{R}^d$ ($j=1,2,\ldots,K$) is the $j$-th cluster center of the referred view in the low $d$-dimensional space. Then, we let:
%We propose a new cluster centers initialization scheme:

\begin{equation}\label{eq:C}
    \pmb{\mu}_j^v=\pmb{c}_j^{V_s}, \forall v\in\{1, 2, \dots, V\}.
\end{equation}

In Eq.~(\ref{eq:q}), the $\pmb{\mu}^v$ of all views are initialized with the same cluster centers. 
Note that the cluster centers of the referred view are shared by other views only in the initialization phase. In the fine-tuning phase, the cluster centers of each view are learned by multi-view collaborative training and only the auxiliary target distribution of the referred view are shared.

The time and space complexities of DEMVC are in the same level with IDEC, which are linear to $N$. This allows our algorithm to deal with large-scale data clustering problems. The proposed DEMVC model is summarized in Algorithm~\ref{alg:alg1}. 

%--------------------------------------------------------------------
%-----------------------------------------------------------------

\begin{algorithm}[htbp]
\caption{Deep Embedded Multi-view Clustering with Collaborative Training} 
\label{alg:alg1}
\begin{algorithmic}[1]
\REQUIRE ~~\\
Multi-view dataset, number of clusters $K$\\
\ENSURE ~~\\
Multi-view cluster assignment $\pmb{s}$\\
Reconstructed samples $\hat{\pmb{x}}^v$
\STATE //\textit{Initialization phase}
\STATE Pre-train deep autoencoders by Eq.~(\ref{eq:Lr})
\STATE Initialize cluster centers by Eq.~(\ref{eq:C})\\
\STATE //\textit{Fine-tuning phase}
\WHILE{not reaching the maximum iterations}
\FOR{$V_r$ in $\{1,2,\dots,V\}$}
\STATE Calculate multi-view prediction ${Q^v}$ by Eq.~(\ref{eq:q})
\STATE Calculate ${P^r}$ of the referred view $V_r$ by Eq.~(\ref{eq:p})
\STATE Fine-tune all the deep autoencoders by Eq.~(\ref{eq:LOSS})
\ENDFOR
\ENDWHILE
\STATE Output $\pmb{s}$ and $\hat{\pmb{x}}^v$ calculated by Eq.~(\ref{eq:s}) and Eq.~(\ref{eq:decoder})
\end{algorithmic}
\end{algorithm}

\section{Experimental Setup}\label{sec:experiments}
\subsection{Datasets}\label{sec:Datasets}

\textbf{NosiyMnist-RotatingMnist (Noisy-Rotating)}.
The Mnist dataset~\cite{lecun1998gradient} collects 70,000 samples of 28$\times$28 pixel size from 10 classes, i.e., digits 0-9. 
Following~\cite{wang2015deep}, we construct RotatingMnist by randomly rotating the images with angles uniformly sampled from $[$-$\pi/4, \pi/4]$. To build NosiyMnist, for each sample in RotatingMnist, we randomly select an image with the same label from the Mnist dataset. Then, each pixel is masked with independent random noise uniformly sampled from $[0, 255]$. After that, the values of all pixels are truncated to $[0, 255]$. In multi-view clustering, NosiyMnist and RotatingMnist are two different views corresponding to each other.

\textbf{Mnist-USPS}. USPS is also a handwritten digital dataset, each sample of which is a 16$\times$16 image. Mnist and USPS are treated as two different views of digits. We use the same dataset as \cite{peng2019comic} did, each view of which contains 5000 digits. Every class provides 500 samples.
%+++++++++++++++++++

\textbf{Fashion-10K}. We use the test set of Fashion dataset~\cite{xiao2017fashion}, which consists of 10,000 28$\times$28 gray images. It contains 10 categories, such as T-shirt, Dress, Coat and is a more challenging dataset. 
We consider this test set as the first view, and then for each sample, we randomly select a sample with the same label from this set to construct the second/third view. Different views of each sample are different individuals from the same category. 

%For the sake of comparison, we note two views of Fashion-10K as Fashion-10K(2 views), note three views of Fashion-10K as Fashion-10K(3 views).

\textbf{Mnist-10K}. The test set of Mnist dataset is used as the first view. The second/third views are constructed in the same way as Fashion-10K.
%Just as we constructed Fashion-10K, based on Mnist-test we construct Mnist-10K. 

The input features of each dataset are scaled to $[0, 1]$.

\subsection{Comparing Methods}
\label{sec:Comparison}
We compare our DEMVC against the following multi-view clustering methods on NosiyMnist-RotatingMnist and Mnist-USPS: 

(1) deep canonical correlation analysis (DCCA)~\cite{andrew2013deep}. 

(2) deep canonically correlated autoencoders (DCCAE)~\cite{wang2015deep}. 

(3) diversity-induced multi-view subspace clustering (DiMSC)~\cite{cao2015diversity}.

(4) latent multi-view subspace clustering (LMSC)~\cite{zhang2017latent}.

(5) binary multi-view clustering (BMVC)~\cite{zhang2018binary}.

(6) multi-view clustering without parameter selection (COMIC)~\cite{peng2019comic}.

In order to illustrate the significant improvement of our DEMVC compared to single-view deep clustering approaches, we test several state-of-the-art deep clustering methods on Fashion-10K and Mnist-10K: 

(1) deep embedded clustering (DEC)~\cite{xie2016unsupervised}.

(2) improved deep embedded clustering (IDEC)~\cite{guo2017improved}.

(3) deep embedded clustering with data augmentation (DEC-DA)~\cite{guo2018deep}.

(4) deep clustering network (DCN)~\cite{yang2017towards}.

(5) $k$-subspace clustering network ($k$-SCN)~\cite{zhang2018scalable}.

(6) neural collaborative subspace clustering (NCSC)~\cite{zhang2019neural}. 

\subsection{Implementation Details}
\label{sec:Details}
\textbf{Network settings}. 
All the used autoencoders are the same convolutional autoencoder. 
Following deep single-view clustering methods \cite{guo2018deep,ren2018deep}, the structure of encoder is: $\text{Input}\rightarrow\text{Conv}_{32}^{5}\rightarrow\text{Conv}_{64}^{5}\rightarrow\text{Conv}_{128}^{3}\rightarrow\text{Fc}_{10}$. That is, the convolution kernel sizes are 5-5-3 and stride of 2 as default, and channels are 32-64-128. The dimensionality is reduced to 10 since the embedded layer $\text{Fc}_{10}$ is a fully connected network, which is made up of 10 neurons.
The encoders and decoders of multiple views are symmetric correspondingly. 
%As in \cite{guo2017improved,guo2018deep,ren2018deep}, 
The ReLU is always chosen as the activation function except for the input, embedded, output and clustering layers. 
%\textcolor{blue}{As shown in Figure \ref{fig:framework}, the input layers and output layers directly accept inputs and form outputs. The activation function of the embedded layers is linear. The clustering layers take the embedded feature output of the embedded layers as input and the clustering centers as the learnable network parameters. The output of the clustering layer is used as the clustering soft label, and the auxiliary target of the referred view is used as the update objective.}

%+++++++++++++++++++++
%\textcolor{blue}{The choice of the first referred view depends on specific dataset. Different views have different difficulty in clustering. Empirically, we chose the view that is easier to cluster as the first referred view. For example, as shown in Table \ref{tab:table3} in Section \ref{sec:Module analysis}, the RotatingMnist clustering performance was significantly lower that of NoisyMnist without collaborative training. So let RotatingMnist be the first referred view is not a good guide and we chose NoisyMnist to be the first referred view to start the algorithm.}

We use Adam and default parameters in Keras$^1$ to optimize the entire networks. The autoencoders of DEMVC are pre-trained for 500 epochs for each view. In the fine-tuning phase, when a view becomes the referred view, it guides each view (including itself) to train 200 batches in an end-to-end manner. The batch size is 256 and the number of fine-tuning iterations is 20,000. 
%USPS is a 16$\times$16 grayscale dataset. For the sake of generality, we make the size become 28$\times$28 by expanding around it with zero.
All experiments of DEMVC are performed on Windows PC with Intel (R) Core (TM) i5-9400F CPU @ 2.90GHz, 16.0GB RAM, and GeForce RTX 2060 GPU (6GB caches). 

\footnote{$^1$\url{https://github. com/fchollet/keras}}

\subsection{Evaluation Measures}
\label{sec:metrics}
The quantitative metrics are adjusted rand index (ARI), unsupervised clustering accuracy (ACC), and normalized mutual information (NMI). A larger value of ARI/ACC/NMI indicates a better clustering result. 
All the reported results (except for those values excerpted from the papers) are the average values of 5 independent runs.

\section{Results and Analysis}
\label{sec:Algorithm Results}

\subsection{Results on Real Data}
\label{sec:results}

\begin{table}
\caption{Comparison of multi-view clustering algorithms. The suffixes `-V1' and `-V2' represent the clustering results of the first view (NoisyMnist or Mnist), and the second view (RotatingMnist or USPS), repsectively.}
    \begin{center}
        \selectfont
 % \caption{Comparison of multi-view clustering algorithms.}
        \begin{tabular}{|c|l|l|l|l|}
            \hline
            \multirow{2}{*}{Methods}&
            \multicolumn{2}{c|}{Noisy-Rotating}&\multicolumn{2}{c|}{Mnist-USPS}\cr\cline{2-5} 
            & ACC & NMI & ACC & NMI\cr
            \hline
            \hline
            DCCA (ICML 2013) & 97.00$^\dagger$  & 92.00$^\dagger$   & 97.42$^*$ & 93.60$^*$\\
            DCCAE (ICML 2015) & 97.50$^\dagger$  & 93.40$^\dagger$  & 98.00$^*$ & 94.70$^*$\\
            DiMSC (CVPR 2015) & \quad-  & \quad-   & 48.34$^*$ & 36.02$^*$\\
            LMSC (CVPR 2017) & \quad-  & \quad- & 78.60$^*$ & 78.49$^*$\\
       % DMSC(J-STSP'18) & -  & -   & 90.11 & 87.19\\
            BMVC (TPAMI 2018) &85.61 &81.48 &88.68 &89.93\\
            COMIC+SC (ICML 2019) & \quad-  & \quad- & 97.44$^*$ & 94.83$^*$\\
            \textbf{DEMVC-V1} &\textbf{99.51}&\textbf{98.37} &\textbf{99.71} &\textbf{99.24}\\
            \textbf{DEMVC-V2} &\textbf{99.08}&\textbf{97.16} &\textbf{99.81} &\textbf{99.35}\\
            \textbf{DEMVC}    &\textbf{99.87}&\textbf{99.53} &\textbf{99.83} &\textbf{99.49}\\
            \hline
        \end{tabular}
    \end{center}
   % \tabnote{}
\label{tab:table1}
\end{table}

\begin{table}
\caption{Comparison of single-view deep clustering algorithms. The suffixes `-2 views' and `-3 views' indicate that 2 and 3 views are built with the corresponding dataset (Fashion-10K and Mnist-10K), respectively.}
  \begin{center}
  \selectfont
    \begin{tabular}{|c|l|l|l|l|}
    \hline
    \multirow{2}{*}{Methods}&
    \multicolumn{2}{c|}{Mnist-10K}&\multicolumn{2}{c|}{Fashion-10K}\cr\cline{2-5} &ACC&NMI&ACC&NMI\cr
    \hline
    \hline
    DEC (ICML 2016) & 83.41   & 79.22 & 56.70 & 61.29\\
    IDEC (IJCAI 2017) & 84.25  & 82.77    & 57.43 & 61.55\\
    DCN (ICML 2017) & 83.31$^\ddagger$  & 80.86$^\ddagger$  & 58.67$^\ddagger$ & 59.40$^\ddagger$\\
    DEC-DA (ACML 2018) & 97.93   & 95.81     &53.55 &59.91\\
    $k$-SCN (ACCV 2018) & 87.14$^\ddagger$ & 78.15$^\ddagger$    & 63.78$^\ddagger$ & 62.04$^\ddagger$\\
    NCSC (arXiv 2019) & 94.09$^\ddagger$ & 86.12$^\ddagger$    & 72.14$^\ddagger$ & 68.60$^\ddagger$\\
    \textbf{DEMVC-2 views} &\textbf{99.87}&\textbf{99.60} &\textbf{84.75} &\textbf{87.14}\\
    \textbf{DEMVC-3 views} &\textbf{99.99}&\textbf{99.96} &\textbf{78.99} &\textbf{90.88}\\
    \hline
    \end{tabular}
    \end{center}
    \label{tab:table2}
\end{table}

Table \ref{tab:table1} shows the results of comparing methods on NoisyMnist-RotatingMnist and Mnist-USPS. Here, the results marked with `${*}$' and `${\dagger}$' are excerpted from \cite{peng2019comic} and \cite{wang2015deep}, respectively.
`-’ denotes that the corresponding methods, which are based on subspace clustering or spectral clustering, are with high complexity and can not solve the clustering task of 70,000 data samples.  
%`-’ denotes that the corresponding methods are with high complexity and can not solve the clustering task of 70,000 data samples~\cite{zhang2019neural}. 
The best three results in each column are highlighted in boldface. It can be seen that DEMVC can always achieve the best performance.
After sufficient fine-tuning, the soft cluster assignments $\pmb{q}_i^1$ and $\pmb{q}_i^2$ are almost the same. In this case, some mispredicted assignments can be corrected to some extent by averaging them. Therefore, DEMVC's final clustering performance is slightly better than that of the two views (DEMVC-V1 and DEMVC-V2). It is worth noting that the aligned degree of multi-view clustering soft assignments can reflect the progress of collaborative training. When the soft clustering assignments of multiple views are almost consistent, it indicates that DEMVC reaches the consensus principle
%in the clustering interpretation of samples 
and meets the stopping criterion.
%of convergence.}
%It can be seen that DEMVC can always achieve the best performance. For DEMVC, the soft cluster assignments $\pmb{q}_i^1$ and $\pmb{q}_i^2$ are almost identical. By averaging them, some wrong assignments can be avoided. Thus, the performance of DEMVC is slightly better than that of the two views. 
%, validating the analysis in Section 3.1.

\begin{figure}
\centering
\subfigure{\includegraphics[height=3.5in, width=6.22in]{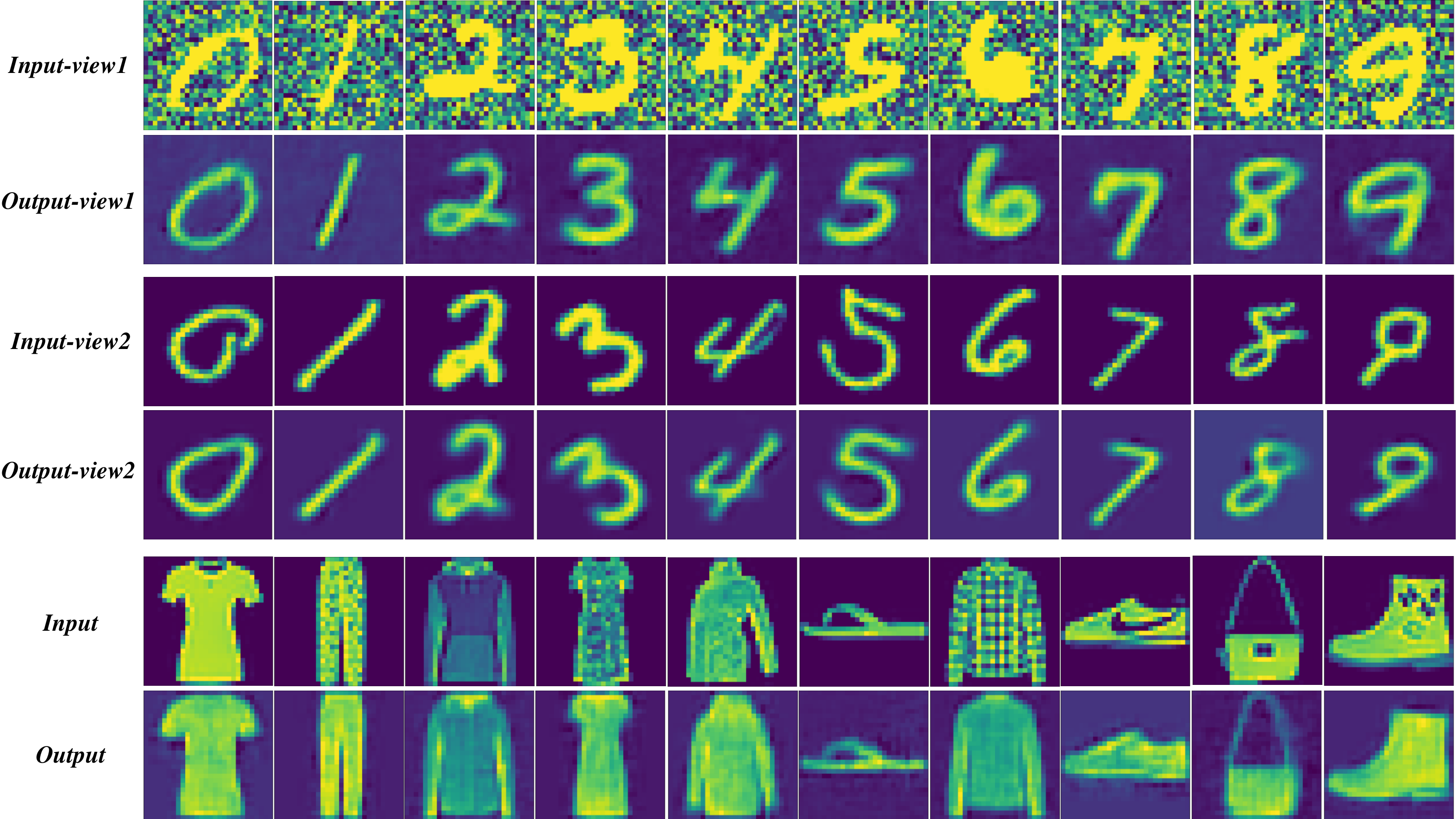}}
\caption{Reconstructed results of multi-view decoders. On NoisyMnist-RotatingMnist, the first and third rows are the corresponding input images of view1 and view2, and the second and fourth rows are the images reconstructed by the corresponding decoders. The fifth and sixth rows are the corresponding input and output images on Fashion-10K. The decoders of DEMVC remove unnecessary noises from the input pictures and make samples look like more standard, indicating their good reconstructed capability.}
\label{fig:Reconstruction}
\end{figure}

\begin{figure}
\centering
\subfigure[NoisyMnist]{\includegraphics[height=2.6in, width=3in]{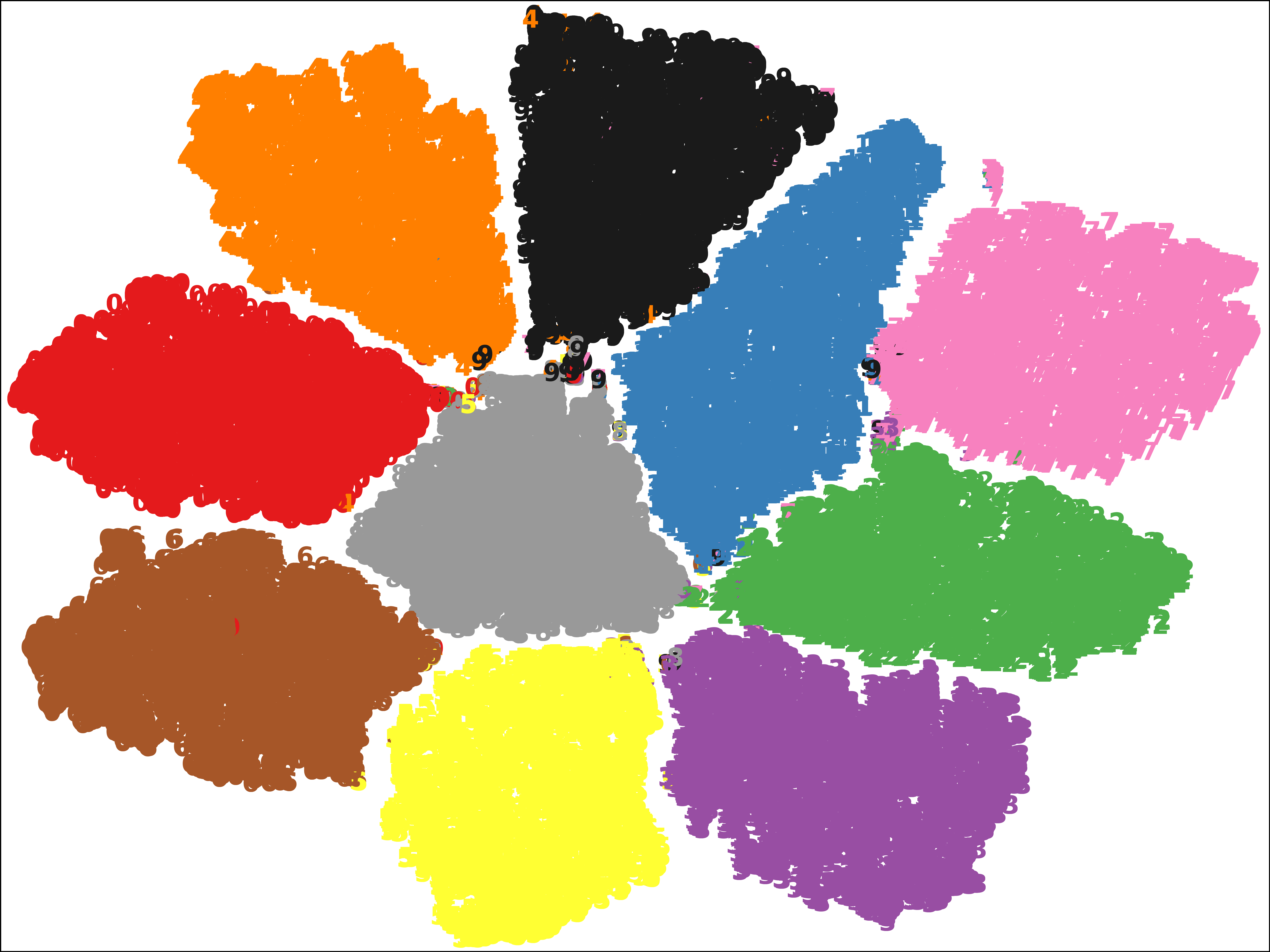}}
\subfigure[RotatingMnist]{\includegraphics[height=2.6in, width=3in]{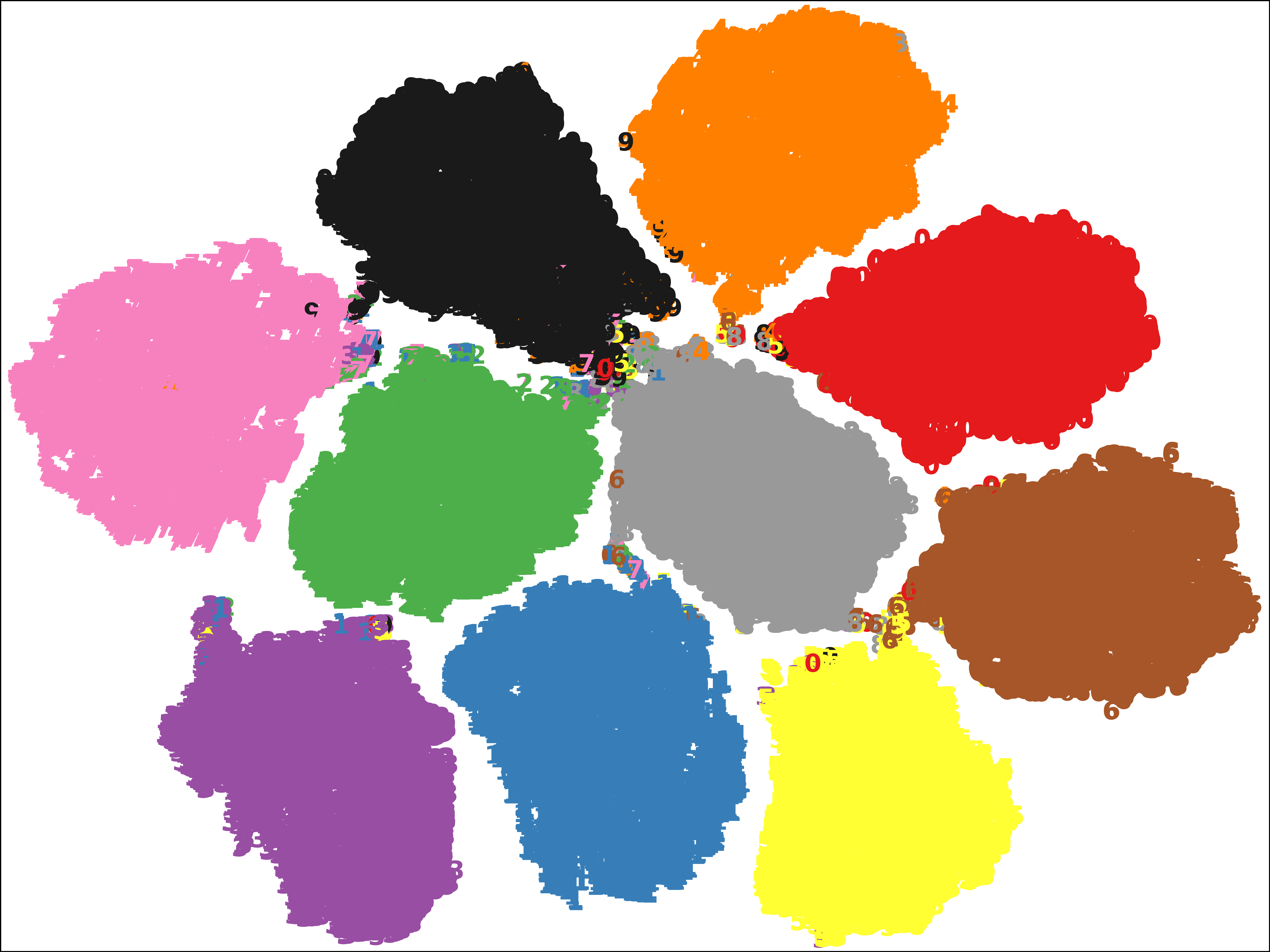}}
\subfigure[USPS]{\includegraphics[height=2.6in, width=3in]{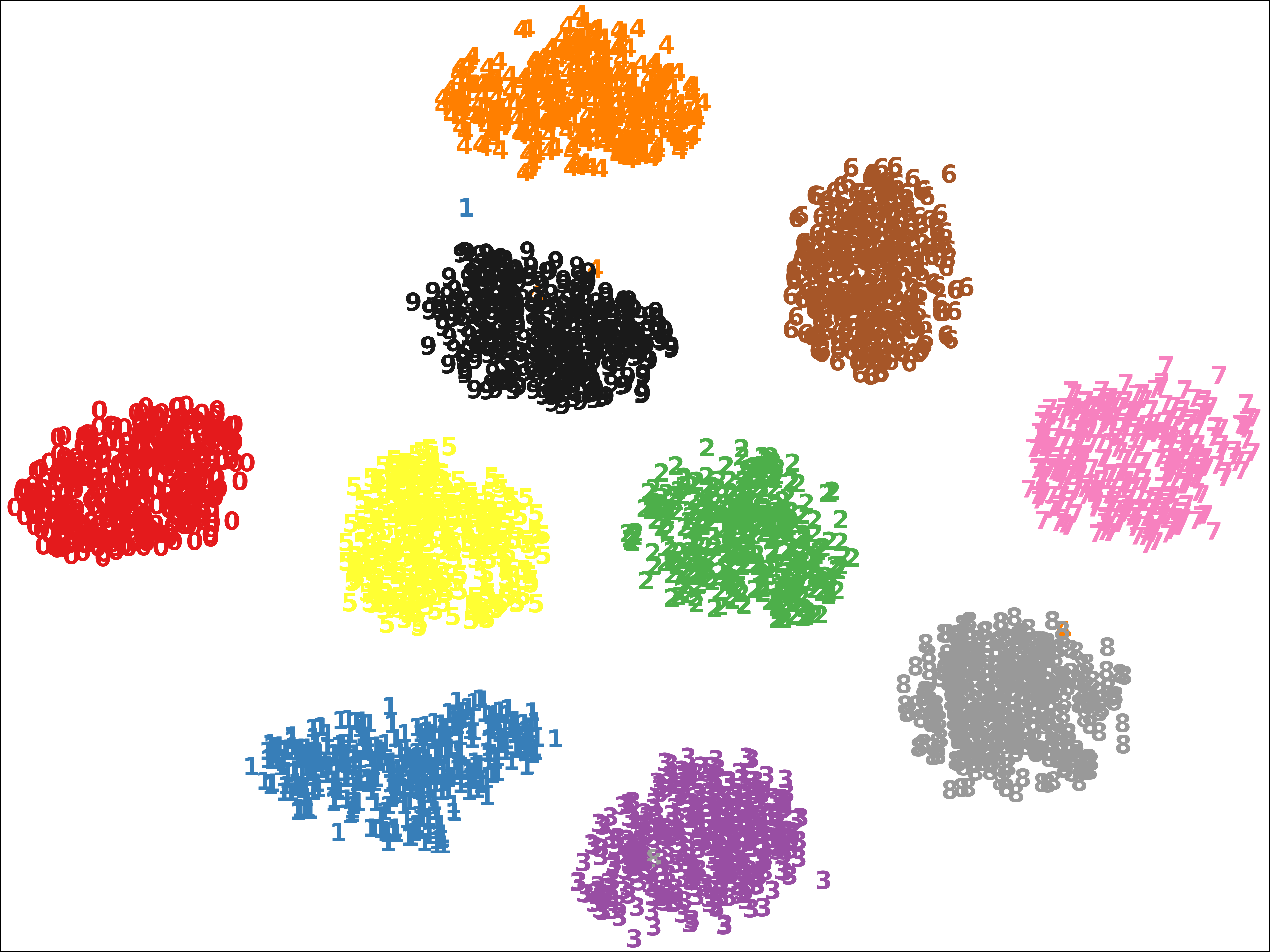}}
\subfigure[Fashion-10K]{\includegraphics[height=2.6in, width=3in]{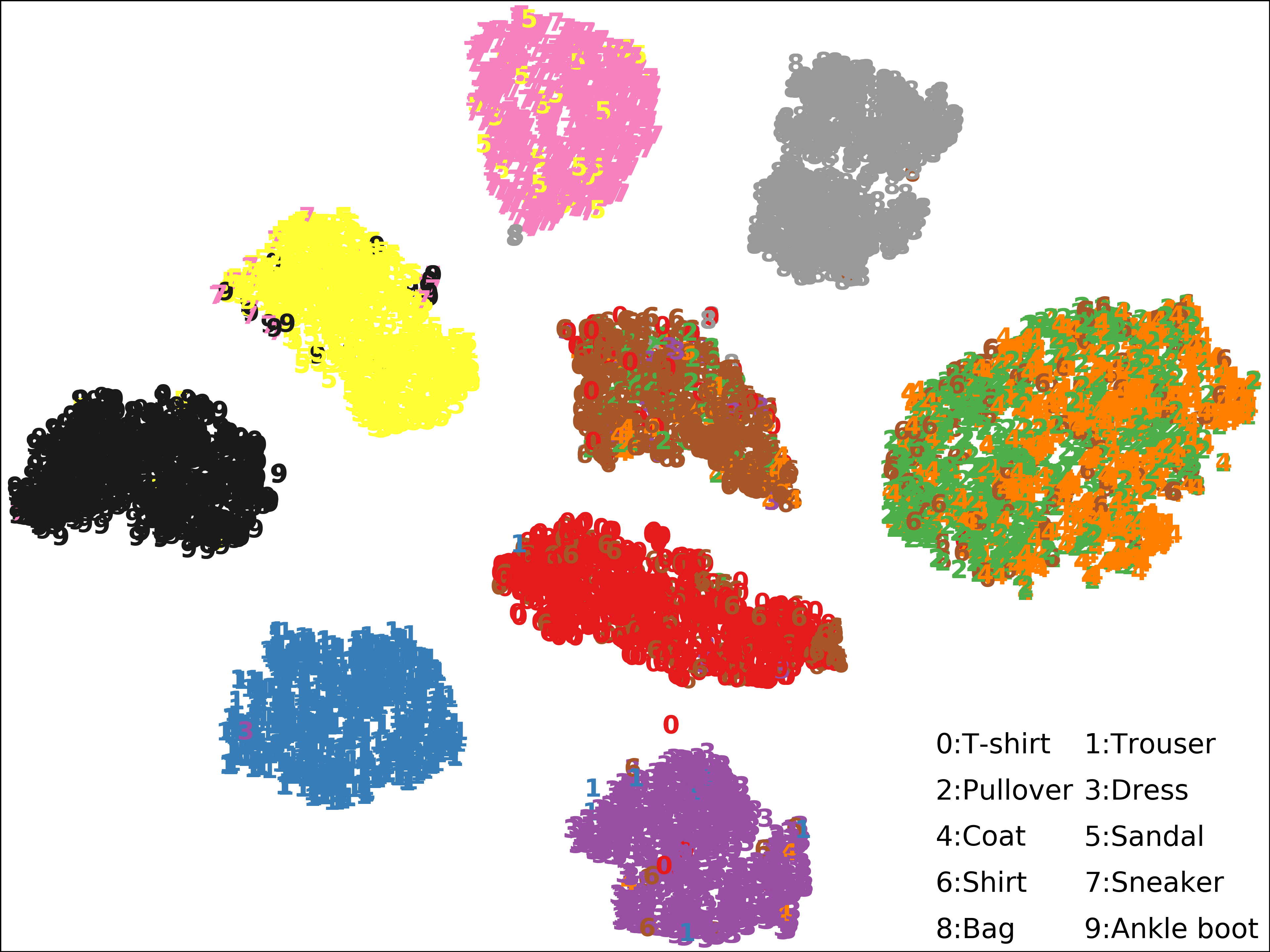}}
\caption{Visualization of the embedded features via $t$-SNE.}
\label{fig:separation}
\end{figure}

When comparing with single-view deep clustering methods, we directly apply them on the test set of Mnist and Fashion because they can only deal with one view clustering task. The results are shown in Table \ref{tab:table2}, where the values with `${\ddagger}$' are obtained from~\cite{zhang2019neural}. The best two values in each column are highligted. In the case of two views, the performance of DEMVC is better than all the single-view methods. The performance of DEMVC with three views is better than that with two views in general. This shows that our approach can effectively extract useful feature from multiple views, and also validates the effectiveness of our approach applies in multi-view clustering.

\subsection{Visualization of Results}
The reconstructed results of DEMVC are shown in Figure \ref{fig:Reconstruction}. 
Specifically, the image ``6'', in the NoisyMnist dataset in the first line, fills the empty space in the hollow position of the digital ``6''. After the sample is reconstructed by the decoder, the redundant part of the hollow position of the image ``6'' is discarded. The handwritten digit ``6'' has a hollow part, which is the common feature of the images ``6'' in the training dataset. The autoencoder obtained by DEMVC accurately captures this common information. In addition, the processing of Fashion-10K by the autoencoder focuses on the contour of the extracted object. The reason is that the similarity of clothing products in Fashion-10K mainly lies in their appearance and shape, while the logo or pattern on the clothing products are not so important. 

It is shown that, on the noisy digits, rotating digits, and fashionable products, DEMVC's multi-view decoders can effectively reconstruct the images based on the low-dimensional embedded features of the input samples. The model can even patch up the missing parts while ignoring the unnecessary information of the samples --- such as noise, rotation and redundant parts --- and finally make the samples look more standard. This indicates its good representation capability of sample features and reconstruction capability, which is the premise to improve clustering performance.
%The premise that the decoders have the function of reconstruction and restoration is that the encoders have good representation capability in low-dimensional space.
%can accurately excavate the essential information in the image samples and compress the information into the low-dimensional space to the greatest extent. 
%The training process of DEMVC's autoencoders is actually representation learning for data samples, and their excellent representation capability is the premise to improve subsequent clustering performance.}

We further visualize the embedded features of four datasets in 2-dimensional space via $t$-SNE~\cite{maaten2008visualizing}, as shown in Figure \ref{fig:separation}. 
Points of the same class are plotted with the same color. For the sake of demonstration, the ground-truth label of each point is also plotted.
We can see that points from the same cluster are highly concentrated and the different clusters are well separated, verifying again the impressive representation capability and clustering performance of DEMVC.
%that the embedded features of DEMVC have impressive representation capability.}

On Fashion-10K, some fashionable products, e.g., pullover, coat, and shirt, look similar in images of 28$\times$28 pixel size. So some points from these classes are closed to each other, as shown in Figure \ref{fig:separation}(d). Nevertheless, the clustering performance of DEMVC is still significantly better than that of other single-view algorithms. We think that more information (e.g, semi-supervised information) is needed to further separate those samples.
%For comparability and fairness, the autoencoders we used are consistent with other methods and their structure are relatively simple, so they are difficult to capture the difference of pixel level. 
%This is why the clustering performance on Fashion-10K is not like that of on Mnist datasets.}
%has reached the upper limit.}

\subsection{Module Analysis}
\label{sec:Module analysis}

\begin{figure}
\centering
\subfigure[IDEC $vs.$ Coo]{\includegraphics[height=2.47in, width=3in]{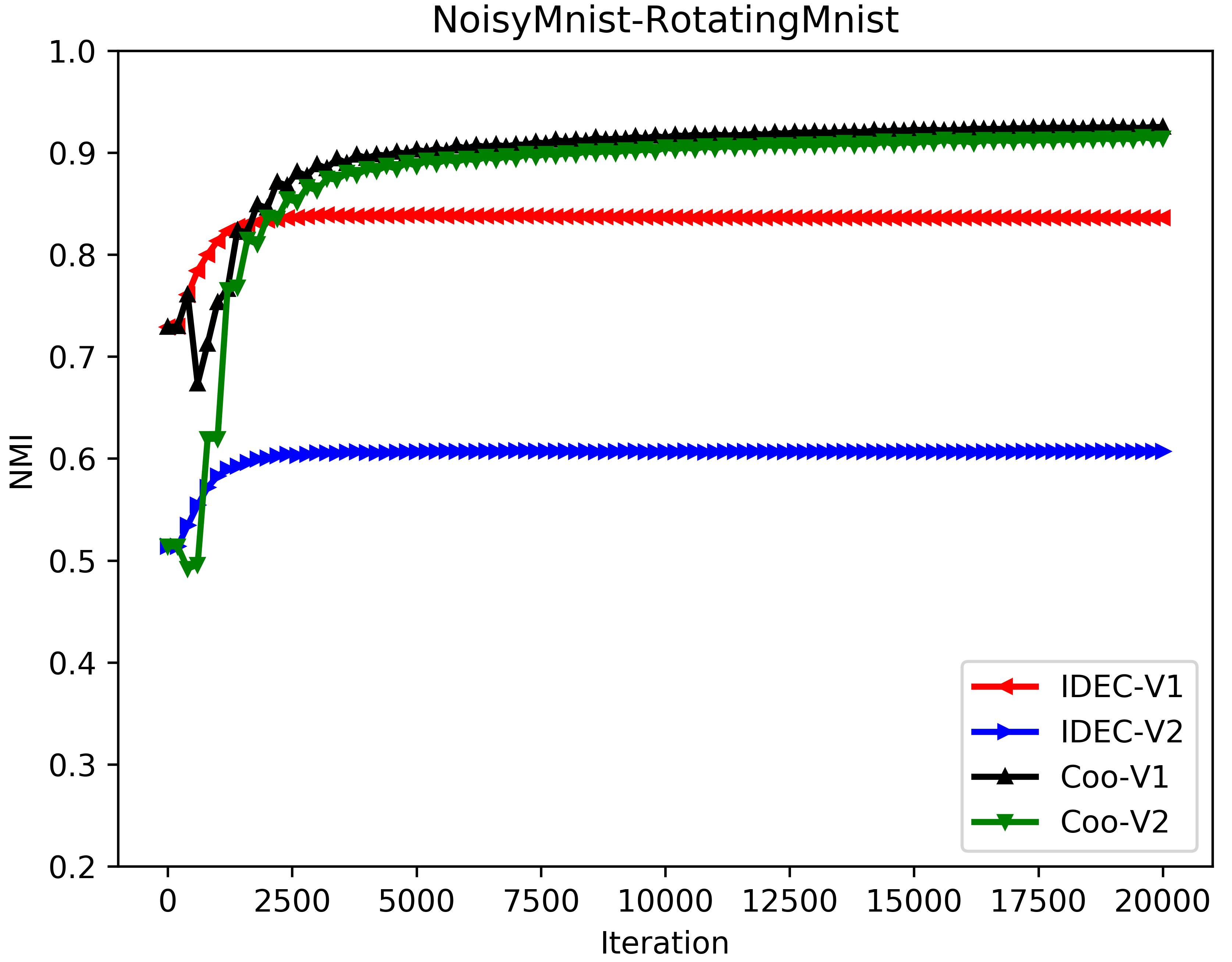}}
\subfigure[IDEC $vs.$ Coo]{\includegraphics[height=2.47in, width=3in]{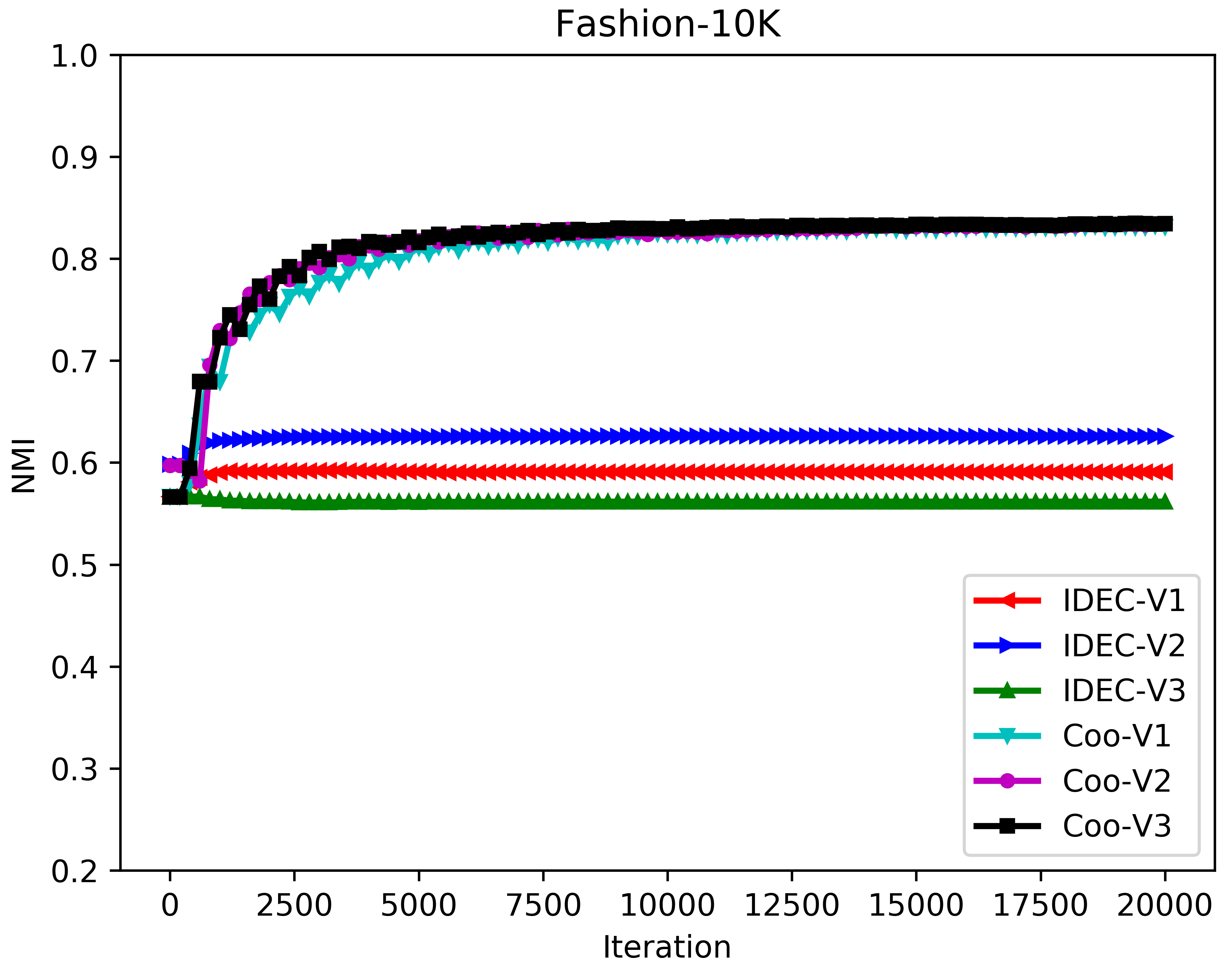}}
\subfigure[Coo $vs.$ Coo+SetC]{\includegraphics[height=2.47in, width=3in]{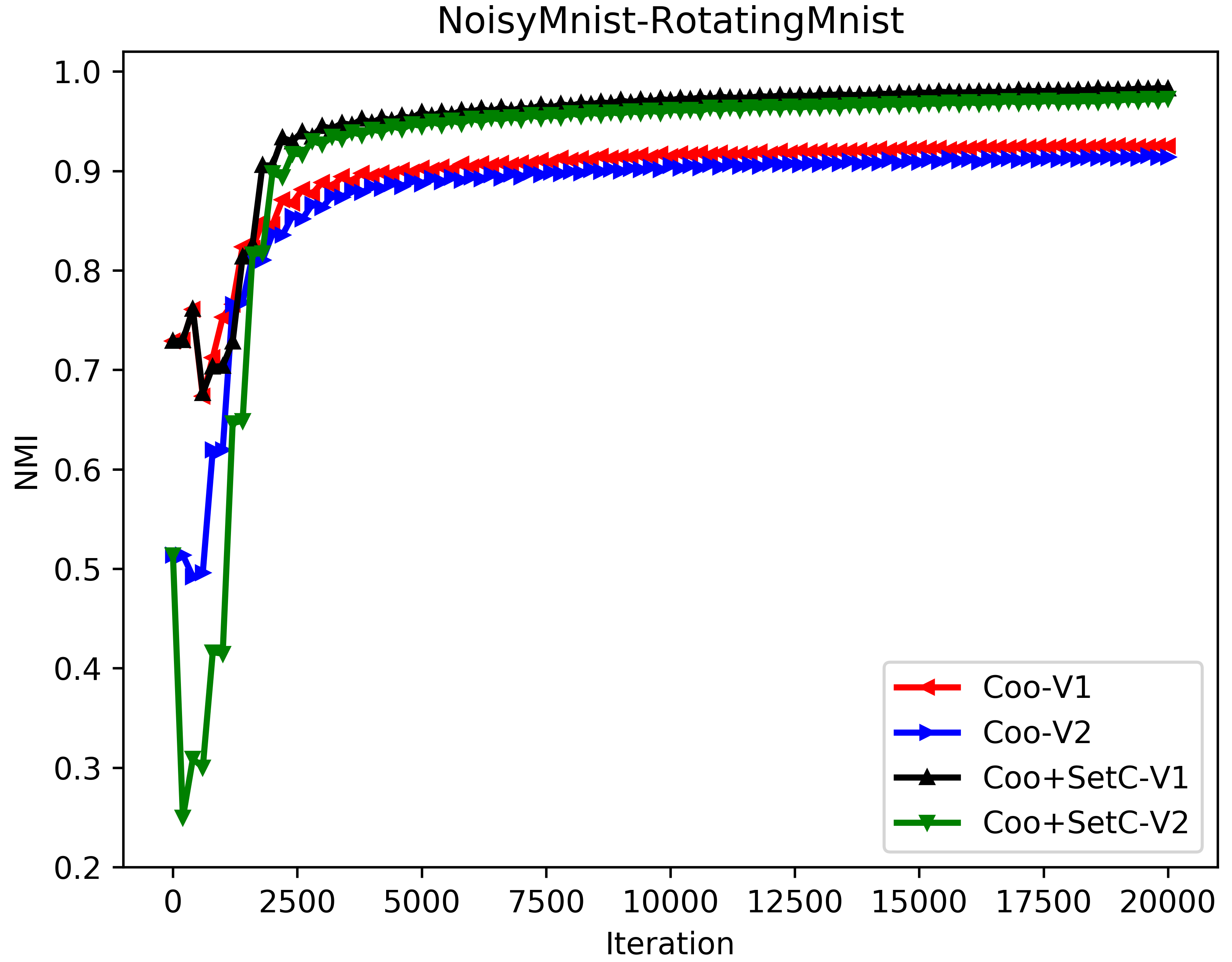}}
\subfigure[Coo $vs.$ Coo+SetC]{\includegraphics[height=2.47in, width=3in]{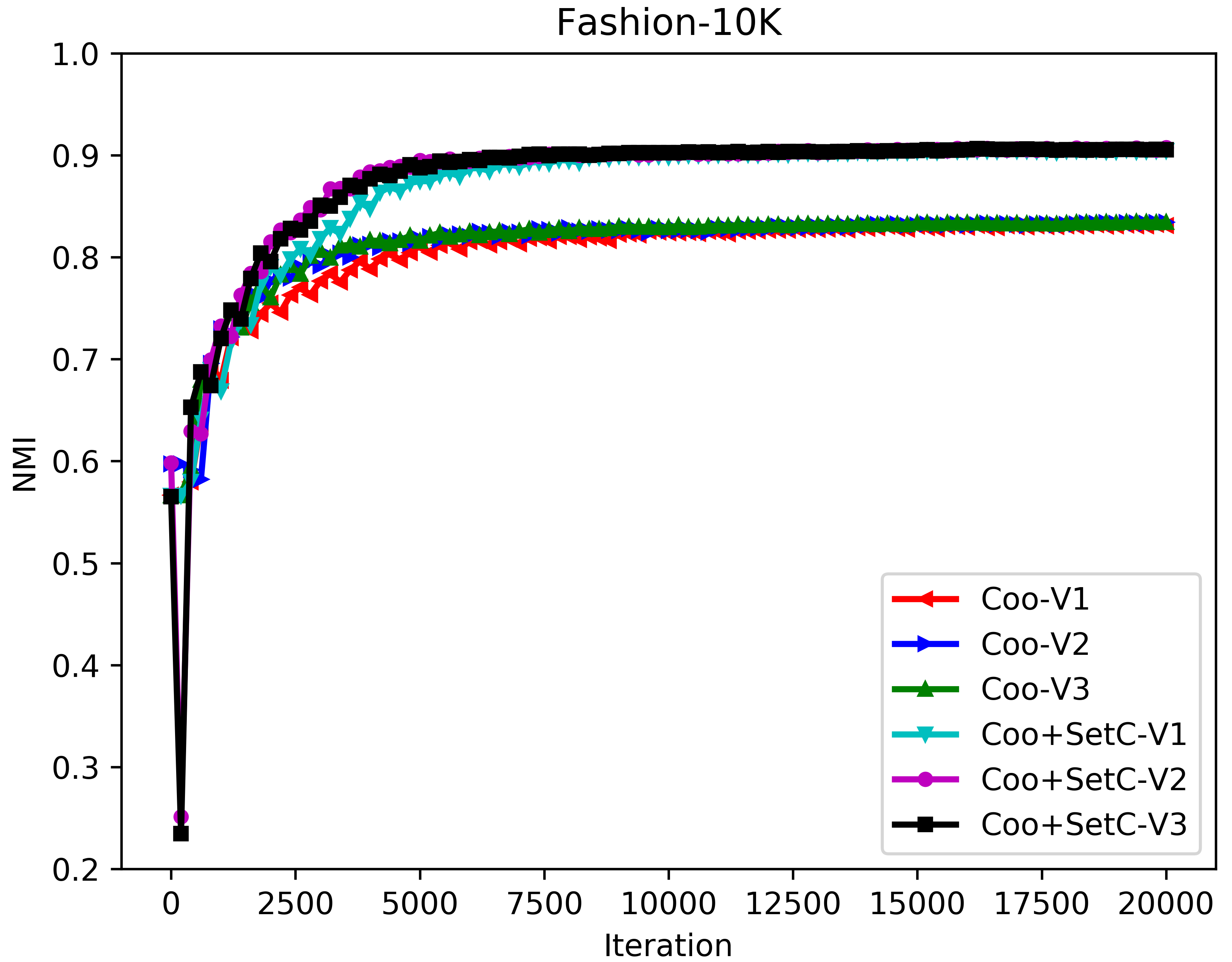}}
\caption{The training process and the comparison of different modules.}
\label{fig:Module}
\end{figure}

\begin{table}
\renewcommand\tabcolsep{4.3pt} 
\centering
\caption{The influence of different modules of DEMVC. `2 views' means two views are constructed for the corresponding dataset.}
\begin{threeparttable}
    \begin{tabular}{ccccccccccccc}
    \toprule
    &\multicolumn{3}{c}{NoisyMnist-RotatingMnist} &\multicolumn{3}{c}{Mnist-USPS} &\multicolumn{3}{c}{Mnist-10K(2 views)} &\multicolumn{3}{c}{Fashion-10K(2 views)}\cr
    \hline
    Module & ACC & NMI & ARI & ACC & NMI & ARI & ACC & NMI & ARI & ACC & NMI & ARI\\
    \hline
    IDEC-V1     &88.91	&83.54	&79.67	&81.24	&78.17	&71.36	&84.25	&82.77	&77.56	&57.43	&61.55	&45.31\\
    IDEC-V2     &59.22	&60.81	&44.05	&66.20	&68.75	&55.72	&83.58	&81.52	&76.52	&48.84	&60.57	&40.82\\
    Coo-V1      &88.53	&93.46	&86.51	&99.64	&98.98	&99.20	&90.09	&96.64	&89.69	&70.21	&81.37	&64.10\\
    Coo-V2      &87.53	&91.76	&84.59	&99.72	&99.23	&99.38	&90.05	&96.54	&89.63 	&70.18	&81.25	&64.10\\
    Coo+SetC-V1 &99.51	&98.37	&98.92	&99.71	&99.24	&99.25	&99.83	&99.49	&99.62 	&84.61	&86.83	&78.49\\
    Coo+SetC-V2 &99.08	&97.16	&97.95	&99.81	&99.35	&99.47	&99.74	&99.28	&99.41 	&84.73	&87.09	&78.73\\
    DEMVC      &99.87	&99.53	&99.71	&99.83	&99.49	&99.63	&99.87	&99.60	&99.70 	&84.75	&87.14	&78.80\\
    \bottomrule
    \end{tabular}
\end{threeparttable}
\label{tab:table3}
\end{table}

This section explores the role of each module of DEMVC. 
In Figure \ref{fig:Module} and Table \ref{tab:table3}, `IDEC-V1' means IDEC is applied on the first view of the corresponding dataset. `Coo' represents collaborative training without the strategy of sharing cluster centers in the beginning. `Coo+SetC' represents collaborative training method with consistency strategy of cluster centers initialization. 

Figure \ref{fig:Module}(a) and Figure \ref{fig:Module}(b) show the results of IDEC and Coo on NoisyMnist-RotatingMnist and Fashion-10K (with three views). The proposed collaborative training constantly switches the referred view, so the multiple views can teach each other such that their clustering performance increases alternately. Overall, for all views, collaborative training significantly outperforms general training, i.e., applying IDEC on each view independently.

Figure \ref{fig:Module}(c) and Figure \ref{fig:Module}(d) give the results of Coo and Coo+SetC. It can be observed that the final clustering performance of Coo+SetC is much better than that of Coo, verifying the effectiveness of the developed consistency strategy of cluster centers initialization. For Coo+SetC, except for the referred view, the NMI of other views is low, because the clustering centers of the referred view may not well represent the clustering centers of these views in the beginning. However, in subsequent training, the performance of all views cooperatively increase benefiting from the collaborative training and consistency strategy and get better results. 
%+++

Results of different modules of DEMVC on four datasets are shown in Table \ref{tab:table3}. This ablation study demonstrating again that the collaborative training and consistency strategy of initial centers are necessary and useful. Ultimately, DEMVC gets more accurate prediction by averaging the soft labels of multiple views. 

\section{Conclusion}
\label{sec:Conclusions}
In this work, we present a novel deep embedded multi-view clustering algorithm (DEMVC). Through multi-view collaborative training, each view can guide all views, in turn, to learn the embedded features. DEMVC also uses a new consistency strategy for cluster centers initialization and follows the consensus and complementary principles of multi-view clustering. The proposed framework can make use of the multi-view common and complementary information to enhance clustering performance. 
%build the autoencoders with strong representation capability. Using the low dimensional embedded features extracted by the encoders, DEMVC can cluster the samples more correctly and repairs the image samples with the decoders. 
Extensive experiments demonstrate the effectiveness of the proposed model. In addition, DEMVC is of $O(N)$ complexity and can be used for large-scale data clustering. 
%has application value for multiple views learning.

\section*{Acknowledgments}
This work was supported in part by the National Key Research and Development Program of China (No. 2018AAA0100204), the National Natural Science Foundation of China (No. 61806043), and the China Postdoctoral Science Foundation (No. 2016M602674).
%---------------------------------
% \printcredits

%% Loading bibliography style file
%\bibliographystyle{model1-num-names}

% \bibliographystyle{cas-model2-names}
\bibliographystyle{unsrt}   
% Loading bibliography database
\bibliography{refs}

\end{document}